# Optimizing Helmet Detection with Hybrid YOLO Pipelines: A Detailed Analysis

Vaikunth M, Dejey D, Vishaal C, and Balamurali S

Department of Computer Science and Engineering, College of Engineering, Guindy, Anna University, Chennai, India

## Abstract

*Helmet detection is crucial for advancing protection levels in public road traffic dynamics. This problem statement translates to an object detection task. Therefore, this paper compares recent You Only Look Once (YOLO) models in the context of helmet detection in terms of reliability and computational load. Specifically, YOLOv8, YOLOv9, and the newly released YOLOv11 have been used. Besides, a modified architectural pipeline that remarkably improves the overall performance has been proposed in this manuscript. This hybridized YOLO model (h-YOLO) has been pitted against the independent models for analysis that proves h-YOLO is preferable for helmet detection over plain YOLO models. The models were tested using a range of standard object detection benchmarks such as recall, precision, and mAP (Mean Average Precision). In addition, training and testing times were recorded to provide the overall scope of the models in a real-time detection scenario.*

## Keywords

*Object Detection, Traffic Safety, YOLO, Deep Learning, Hybrid Architecture, CNN*

## 1. Introduction

In many countries, traffic accidents involving motorbikes and scooters are among the top causes of injury and death. Helmets are widely recognized as one of the most effective ways to prevent fatal or severe head injuries. Helmet detection presents an object detection challenge where models must accurately determine if a rider is wearing a helmet, often in real-time situations. While traditional computer vision techniques have their merits, they also face limitations in terms of processing speed, accuracy, and adaptability to various environments. The emergence of deep learning-based object detection algorithms has transformed this area, particularly with the introduction of the YOLO series of models.

YOLO's capability for fast, real-time object detection makes it particularly suitable for helmet detection applications. This paper examines the performance of YOLOv8, YOLOv9, and YOLOv11, and their hybridized version for identifying helmets on bike and motorbike riders.

The YOLO family of models has widely been applied in various object detection tasks due to its efficiency in balancing speed and accuracy. Initially proposed by Redmon et al. [1], YOLO frames the task of object detection as a single regression problem that significantly reduces the time of detection in comparison to the earlier methods such as Recurrent-Convolutional Neural Network (R-CNN) or Single Shot Detector (SSD). With each consecutive version of YOLO, architects have added improvements that bring benefits in terms of accuracy, reduced latency, or lower computational load.

Indeed, two of the most critical ideas that the present study focuses on are as follows: helmet detection is one of the most important initiatives undertaken to make roads safe, and the automation of helmet detection systems especially those utilized in the traffic monitoring setup. According to the National Safety Council, helmet usage by motorcyclists alone is estimated to be 37% more effective in preventing fatal injuries. In this regard, the model proposed here can be an excellent tool for law enforcement agencies to discover bike riders who do not wear helmets. Such a system could reduce casualties from bike and motorbike accidents.

## 2. LITERATURE SURVEY

YOLO models have evolved substantially since their inception. One key work in this area is a comprehensive review of YOLO architectures, tracing their development from YOLOv1 to YOLOv8. This study highlights the model's single-stage approach, where object localization and classification are performed simultaneously, making it highly efficient for real-time applications such as autonomous vehicles and video surveillance [2]. Another notable study proposes Complexer-YOLO, which integrates 3D object detection into the YOLO framework using semantic point clouds. This extension enhances the model's performance for real-world tasks like autonomous driving by improving detection accuracy in complex environments [3]. Additionally, another study emphasizes the integration of YOLO with a sliding innovation filter for object tracking in dynamic environments, addressing challenges like occlusions and disturbances, and showcasing YOLO's adaptability beyond static image detection [4].

While YOLO models excel in speed, other methods offer a different tradeoff between accuracy and computational complexity. Faster R-CNN, for instance, is a two-stage object detector that significantly improves precision by first generating region proposals and then refining them using convolutional neural networks. Though slower, it remains a top choice for tasks where accuracy is critical, such as medical imaging and precise localization [5][6]. SSD (Single Shot MultiBox Detector), like YOLO, adopts a single-stage approach but focuses on balancing speed with improved accuracy over simpler YOLO versions [7]. Another significant development is RetinaNet, which introduces the focal loss function to address the problem of class imbalance, thereby boosting accuracy in detecting small objects, a common issue in real-world applications [8] L Nkabulo et. al. performed an ensemble of faster R-CNN and YOLO by first passing the input to the YOLO model and then refining the output using faster R-CNN [9] Lastly, Mask R-CNN extends Faster R-CNN to not only detect objects but also perform instance segmentation, making it highly effective in tasks requiring pixel-level precision, such as robotics and autonomous systems [10].

The above survey has been represented in Table 1 contains the previous highly notable works that have inspired this work and are concerned with object detection. Thus the table highlights the key differences between existing methodology and the methodology proposed in this paper.

Table 1. Literature survey of some works concerned with 2-D and 3-D object detection.

| S. No. | Authors | Methodology | Dataset |
|--------|---------|-------------|---------|
| 1 | J. Terven et al. [2] | YOLOv1-v8, YOLO-NAS | PASCAL VOC 2007, 2012, MS COCO |
| 2 | M. Simon et al. [3] | Complexer-YOLO | KITTI |
| 3 | A. Moksyakov, et al. [4] | YOLO in conjunction with estimation theory filters | MS COCO and custom dataset |
| 4 | S. Ren et al. [5] | Faster R-CNN | PASCAL VOC 2007, 2012, MS COCO |
| 5 | J. Fan, J. Lee et al. [6] | Faster R-CNN+YOLOv2 | Custom dataset |
| 6 | W. Liu et al. [7] | Single Shot multibox Detector (SSD) | PASCAL VOC, COCO, and ILSVRC |
| 7 | T.-Y. Ross and G. Dollár [8] | RetinaNet | MS COCO |
| 8 | L. Nkalubo, R. Nakibuule, and N. Okila [9] | YOLOv5+Faster R-CNN, RetinaNet, Cascade R-CNN, SSD | Custom dataset |
| 9 | K. He, G. Gkioxari et al. [10] | Mask R-CNN | MS COCO |
| 10 | M. Vaikunth et al. [This work] | h-YOLOv8, v9, v11 and YOLOv8, v9, v11 | Online [11] and custom dataset |

While the literature is dominant in numbers with independent object detection systems, a deeper study of combined architectures needs to be explored further. Rightly so, this work compares a set of recent YOLO models by manipulating the YOLO pipeline by including a lightweight custom (CNN) that feeds features to the YOLO model for enhanced object detection. Further details as to how the overall pipeline is designed will be discussed in the next section.

## 3. PROPOSED METHODOLOGY

The development of the Helmet Detection System follows a systematic approach designed to ensure high accuracy and robustness. This approach is divided into several key stages, including dataset Collection, preprocessing, image augmentation, model training, hyperparameter tuning, and model testing as described in Figure 1.

### 3.1. Dataset Collection

The process begins with gathering images depicting individuals both wearing and not wearing helmets. The dataset is curated from two main sources: an online database [11] and custom images captured using phone cameras. This variety ensures that the images feature individuals from different perspectives and under varying lighting conditions, which is crucial for building a model that performs reliably in real-world scenarios. The total image count exceeds 3500.

### 3.2. Preprocessing

Once the dataset is collected, preprocessing is carried out to standardize the images and prepare them for model training. This includes converting all images to RGB format for uniformity and resizing them to a consistent resolution. Pixel normalization is also applied to scale the pixel values, which enhances the efficiency of the model training process by ensuring that all input data is on the same scale.

To further improve the robustness of the model, image augmentation is applied. By introducing variations in the dataset through techniques such as rotation, flipping, zooming, and adjusting brightness and contrast, the model is exposed to a wider range of possible image scenarios while simultaneously eliminating the need for the requirement of a large number of training images [12]. This helps prevent overfitting, as the model becomes more adaptable to changes in the real world, such as lighting and angle variations.

### 3.3. Model Training

After splitting the data into training and testing sets, the training phase begins. This phase involves a specific training pipeline that involves the usage of CNNs to feed features into the independent YOLO model. In specific, three CNN models have been used in sequence as shown in Figure 1. The first CNN layer is shown in Figure 2 for reference and the subsequent CNN layers, which are otherwise similar to the first, vary only in the number of filters, dimensions, and padding. From Figure 2, it is observable that after kernel application, batch normalization followed by activation is carried out.

Normalization is a crucial part of creating a robust model since it allows for data of different types to be brought into a common scale for further processing. Hence, it is a very popular preprocessing tool. In the case of a CNN, the hidden layer outputs can be normalized for faster training. The normalization of input coming from another layer is called batch normalization. This process has also proved to bring stability to the CNN training.

### 3.4. Hyperparameter Tuning

Fine-tuning the hyperparameters is a key step in optimizing the model's performance. Important parameters, such as the number of epochs, batch size, learning rate, and choice of optimizer are adjusted to strike a balance between training efficiency and accuracy. Adam Optimizer is chosen for its adaptive learning rate, which speeds up convergence [13][14]. The learning rate is carefully tuned to ensure effective learning without skipping important minima. Likewise, the batch size and number of epochs are adjusted to improve the model's training speed and accuracy.

### 3.5. Model Testing

Once trained, the model undergoes thorough testing to validate its performance. Techniques like image fusion are employed to refine detection accuracy by combining outputs from different images and also a data enhancement method to overcome the shortcomings that high noise offers [15]. Evaluation metrics, such as precision and recall provide insight into classification performance. Additionally, the Mean Average Precision (mAP) score is used with the Intersection over Union (IoU) set to a 50% confidence threshold to assess how well the predicted bounding boxes align with the actual helmet regions in the images.

### 3.6. Individual YOLO Model Description

All in all, six YOLO models have been used – h-YOLOv8, h-YOLOv9, h-YOLOv11, and their corresponding independent versions.

Independent YOLO models have been extensively studied as testified by the literature review in this manuscript. Although independent models deliver satisfactorily and, in some cases, outstanding results, it is important to strive for as much reliability as possible without overfitting the model.

The independent YOLO models have a different architecture compared to hybridized systems. This architecture involves identical steps as the h-YOLO but devoid of the hybridization block as depicted in Figure 1. Everything else from data preprocessing to data testing remains the same. This setup ensures that the comparison between the independent and hybridized versions remains justified as the hybridization block will be the only varying parameter. Detailed analyses of the independent YOLO models are given in the following subsections.

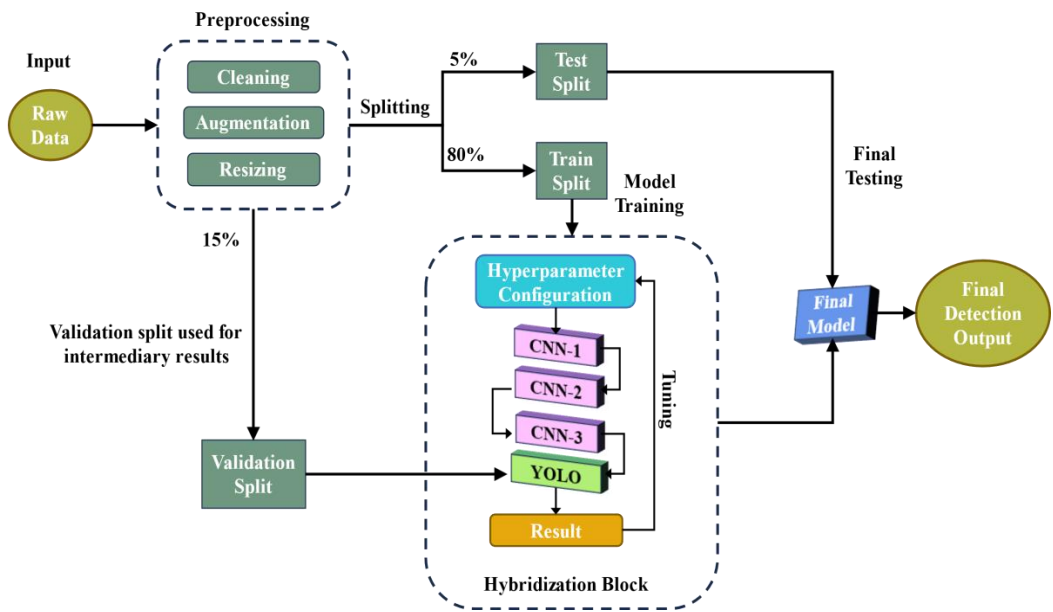

Figure 1. Overall architectural diagram of h-YOLO containing the flow of data from the raw data directory to the final object detection output.

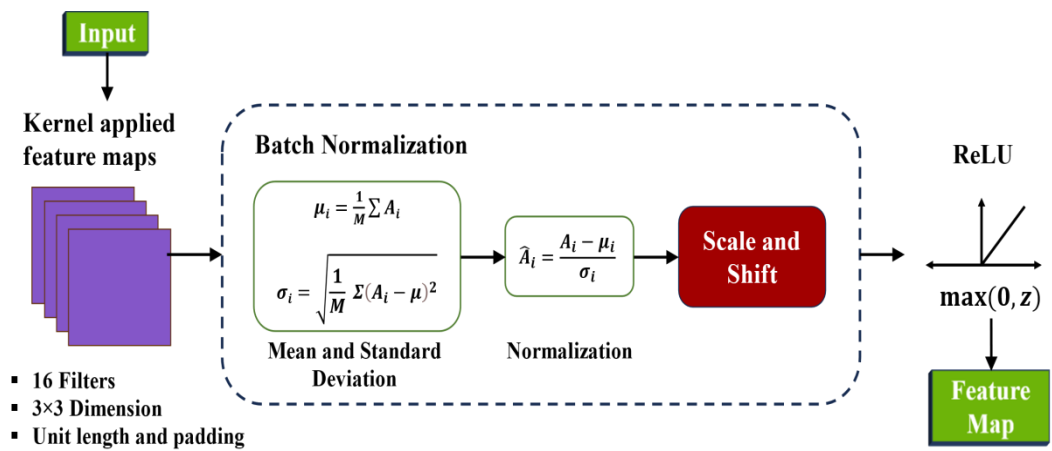

Figure 2. Architectural diagram of the first layer of CNN used in the overall hybrid model.

### 3.6.1. YOLOv8

YOLOv8 focuses heavily on improving multi-object detection accuracy and introducing better handling of small objects within cluttered scenes [16]. Additionally, the model has been optimized for edge devices, which means it can deliver high accuracy without requiring extensive computational resources.

A key innovation in YOLOv8 is the use of attention mechanisms integrated into the network. These mechanisms allow the model to focus on critical regions in the image, leading to better detection in complex scenes.

### 3.6.2. YOLOv9

This model has made significant improvements in terms of global context awareness, which is crucial for tasks like object detection in videos or complex scenes where spatial dependencies matter. YOLOv9 is the first in the series to utilize a hybrid CNN-Transformer backbone, which enhances its ability to capture both local and global features of objects. Advancements such as the introduction of GELAN (Generalized Efficient Layer Aggregation Network) and PGI (Programmable Gradient Information) significantly improve feature extraction and gradient flow capabilities [17].

YOLOv9 also introduces a new loss function that better handles the problem of class imbalance, improving its accuracy for datasets with skewed object distributions. The model also benefits from further improvements in data augmentation techniques and adaptive learning rates, making it more resilient to variations in data quality and scale. As will be described in the results and discussion session, this model proves to be the model with the most mAP score for the utilized data, however with a delay in time taken to achieve said results.

### 3.6.3. YOLOv11

YOLOv11 is the newest iteration of the YOLO series of models. It features improved feature extraction capabilities on its own due to the enhanced backbone and neck architecture [18]. This version introduces boosted deployment capabilities across various platforms like the cloud and systems supporting NVIDIA GPUs.

While each of these YOLO models offers specific improvements over its predecessor, the hybridized approach, which combines the strengths of the pre-CNN, has been employed in this study. Since the CNN is lightweight, in the sense that it does not contain a massively interconnected network of thousands and thousands of neurons, instead it uses neurons in the range of tens and hundreds [19]. This ensures that although the features are being fed to the YOLO model ahead, the pre-CNN does not increase the time taken significantly. However, the features fed by the CNN result in higher recall and mAP scores. Hence the trade-off for a small increment in time taken is justifiable for the substantial increase in model scores.

## 3.7. Testing and Evaluation

The validity of the proposed models has been judged using some of the standard metrics used for the purpose of object detection in literature. In specific, this work utilizes precision, recall, and mAP@50 score. Precision can be defined as the ability of the model to correctly identify positive instances of a particular class. On the other hand, recall measures how many instances of a particular class were correctly detected by the model compared to all the instances of that particular class. Both these metrics are mathematically represented in the equations given below.

$$Precision = \frac{TP}{TP + FP}$$

$$Recal = \frac{TP}{TP + FN}$$

Here, TP refers to true positives, FP refers to false positives and FN refers to false negatives. While experimenting with the hyperparameters, such as the confidence threshold, of the model to achieve different precision and recall scores, it is observed that there is typically a trade-off between precision and recall.

Precision and recall have different purposes. For example, one should aim for a higher precision score when the false negatives are to be tightly controlled while one should aim for a higher recall score when it is important to maximize the identification of a particular class while the presence of false negatives can be tolerated to an extent. In most real-world scenarios though, the optimization of both these scores is crucial and this is where the F1 score that combines both recall and precision comes into play giving the analyzer a comprehensive view of the model's performance. As is seen in the equation below.

$$F1\ score = 2 \times \frac{Precision \times Recall}{Precision + Recall}$$

Another important criterion that is used to criticize a model is the Mean Average Precision (mAP) score. This builds on the average precision (AP) values which is essentially the average of precision scores across a range of recall values and at a particular confidence threshold. When this average is calculated for precision-recall values over a set of confidence thresholds, it is said to be the mean of average precision, hence mAP. In this manuscript, the usage of mAP at a specific Intersection over Union (IoU) value, i.e. 50%, is taken into consideration, and therefore it is mentioned as mAP@50. However, to facilitate detailed analysis and due to a stronger representation of overall performance; mAP has been preferred over F1 in this manuscript. On the contrary, the area under the receiver operating characteristics (AUROC) cannot be considered a justifiable metric for object detection tasks due to its reliance on true negatives (TN). This is because the true negatives in object detection refer to all those bounding boxes that do not contain an object, in this case, a helmet; and therefore TN is practically infinite.

## 4. RESULTS AND DISCUSSION

Evaluation of the various models has been conducted using the rubrics mentioned in the previous section. From Table 1, it is understood that the h-YOLO models categorically produce better precision, recall, and mAP compared to plain YOLO models. When it comes to training and testing time, as shown in Table 2, plain YOLO models take less time than h-YOLO models.

While tackling a safety enforcement task, as is done in this manuscript, it is important to prefer higher scores over faster times not to say that this priority is absolute though, rather, it varies depending on the problem statement. What this means is that one should prefer the model with higher accuracy only when the accuracy difference between the juxtaposed models is significant. Hence, the h-YOLO models are preferred over independent YOLO models as the accuracy difference is 2-3\% which is significant considering the objective of the project can potentially deal with civilian life.

As the comparison between the non-hybridized and the hybridized model has been sorted; naturally, the comparison to find the best among the hybridized systems needs to be conducted.

This can be inferred by analyzing results in Table 1 and Table 2. The h-YOLOv9 model has the best precision, recall, and mAP@50 scores, but h-YOLOv11 only minutely trails while maintaining strikingly shorter training and testing times.

The shorter run times of the h-YOLOv11 models can be explained by analyzing the attributes of the underlying YOLOv11 model. According to the official documentation of the YOLOv11 model, it is stated that the YOLOv11 model has been subjected to enhancement of training and inference pipeline. This enhancement contributes to the overall speed of the system. Moreover, the YOLOv11 model has fewer parameters than the v8 and v9 versions.

Therefore, it is understood that the h-YOLOv11 model provides a reliable balance between detection ability and inference speed making it an ideal choice for real-time detection tasks.

Table 2. Comparison of performance metrics between various models.

| Model | Precision | Recall | mAP@50 |
|---|---|---|---|
| h-YOLOv8 | 0.853 | 0.912 | 0.925 |
| YOLOv8 | 0.792 | 0.91 | 0.905 |
| h-YOLOv9 | 0.887 | 0.9 | 0.932 |
| YOLOv9 | 0.852 | 0.88 | 0.906 |
| h-YOLOv11 | 0.867 | 0.91 | 0.914 |
| YOLOv11 | 0.863 | 0.859 | 0.892 |

Table 3. Training and testing times of different models.

| Model | Training time (min) | Testing time (ms) |
|---|---|---|
| h-YOLOv8 | 78 | 13.8 |
| YOLOv8 | 74 | 10.7 |
| h-YOLOv9 | 134 | 36.4 |
| YOLOv9 | 133 | 43.0 |
| h-YOLOv11 | 101 | 11.9 |
| YOLOv11 | 99 | 10.1 |

## 5. CONCLUSIONS

To encapsulate, helmet detection is an important task in terms of public safety and for technological advancement in automated systems. The YOLO models gained a stronghold as leaders in the arena of real-time object detection. This manuscript is focused on providing a comparative analysis between YOLOv8, YOLOv9, and the latest YOLOv11 and their hybridized models thereby opening up new avenues of thought regarding speed, accuracy, and computational load trade-offs in the context of helmet detection applications. The task of detecting helmets or any other safety equipment such as rear-view mirrors or riding shoes for that matter can have a multitude of applications in enforcing road safety laws. Specifically, such automated detections using machine learning algorithms and frameworks can allow the concerned authorities to take action when due. While the implementation of such technology that can detect defaulters is outside the scope of this paper, the first step of detecting helmets with practical reliability has been achieved, evidently so by the results expressed in the previous section. Therefore, this paper confirms the following --- overall reliability of using machine learning frameworks, the superiority of h-YOLO over independent YOLO by 2-3\% mAP score, and the ability of h-YOLOv11 specifically to balance speed and performance. Future work can involve utilizing h-YOLO to not only detect helmets but also to identify number plates on vehicles. This would allow the concerned authorities to take necessary action and enforce road safety. This manuscript can moreover serve as a guide for the prospective implementation of helmet-detection programmatic chips in traffic cameras.

# REFERENCES

[1] J. Redmon, "You only look once: Unified, real-time object detection," in Proceedings of the IEEE conference on computer vision and pattern recognition, 2016.

[2] J. Terven, D.-M. Córdova-Esparza, and J.-A. Romero-González, "A comprehensive review of yolo architectures in computer vision: From yolov1 to yolov8 and yolo-nas," Machine Learning and Knowledge Extraction, vol. 5, no. 4, pp. 1680–1716, 2023.

[3] M. Simon, K. Amende, A. Kraus, J. Honer, T. Samann, H. Kaulbersch, S. Milz, and H. Michael Gross, "Complexer-yolo: Real-time 3d object detection and tracking on semantic point clouds," in Proceedings of the IEEE/CVF Conference on Computer Vision and Pattern Recognition Workshops, 2019, pp. 0–0.

[4] A. Moksyakov, Y. Wu, S. A. Gadsden, J. Yawney, and M. AlShabi, "Object detection and tracking with yolo and the sliding innovation filter," Sensors, vol. 24, no. 7, p. 2107, 2024.

[5] S. Ren, K. He, R. Girshick, and J. Sun, "Faster r-cnn: Towards real-time object detection with region proposal networks," IEEE transactions on pattern analysis and machine intelligence, vol. 39, no. 6, pp. 1137–1149, 2016.

[6] J. Fan, J. Lee, I. Jung, and Y. Lee, "Improvement of object detection based on faster r-cnn and yolo," in 2021 36th International Technical Conference on Circuits/Systems, Computers and Communications (ITC-CSCC). IEEE, 2021, pp. 1–4.

[7] W. Liu, D. Anguelov, D. Erhan, C. Szegedy, S. Reed, C.-Y. Fu, and A. C. Berg, "SSD: Single shot multibox detector," in Computer Vision–ECCV 2016: 14th European Conference, Amsterdam, The Netherlands, October 11–14, 2016, Proceedings, Part I 14. Springer, 2016, pp. 21–37.

[8] T.-Y. Ross and G. Dollár, "Focal loss for dense object detection," in proceedings of the IEEE conference on computer vision and pattern recognition, 2017, pp. 2980–2988.

[9] L. Nkalubo, R. Nakibuule, and N. Okila, "Real-time object detection using an ensemble of one stage and two stage object detection models with dynamic fine-tuning using kullback-leibler divergence," Authorea Preprints, 2023. [Online]. Available: https://doi.org/10.22541/au.168788163.30701797/v1

[10] K. He, G. Gkioxari, P. Dollár, and R. Girshick, "Mask r-cnn," in Proceedings of the IEEE international conference on computer vision, 2017, pp. 2961–2969.

[11] B. Helmets, "Bike helmet detection dataset," https://universe.roboflow.com/bike-helmets/bike-helmet-detection-2vdjo, sep 2021, visited on 2024-11-09. [Online]. Available: https://universe.roboflow.com/bike-helmets/bike-helmet-detection-2vdjo

[12] D. M. Montserrat, Q. Lin, J. Allebach, and E. J. Delp, "Training object detection and recognition cnn models using data augmentation," Electronic Imaging, vol. 29, pp. 27–36, 2017.

[13] C. Meng, S. Liu, and Y. Yang, "Dmac-yolo: A high-precision yolo v5s object detection model with a novel optimizer," in 2024 International Joint Conference on Neural Networks (IJCNN). IEEE, 2024, pp. 1–8.

[14] P. K. Rajput, K. K. Ravulakollu, N. Jagadam, and P. Singh, "Adam optimizer based deep learning approach for improving efficiency in license plate recognition," in Automation and Computation. CRC Press, 2023, pp. 439–449.

[15] S. Li, Y. Li, Y. Li, M. Li, and X. Xu, "Yolo-firi: Improved yolov5 for infrared image object detection," IEEE access, vol. 9, pp. 141 861–141 875, 2021.

[16] B. Khalili and A. W. Smyth, "Sod-yolov8—enhancing yolov8 for small object detection in aerial imagery and traffic scenes," Sensors, vol. 24, no. 19, p. 6209, 2024.

[17] M. Yaseen, "What is yolov9: An in-depth exploration of the internal features of the next-generation object detector," arXiv preprint arXiv:2409.07813, 2024.

[18] M. A. R. Alif, "Yolov11 for vehicle detection: Advancements, performance, and applications in intelligent transportation systems," arXiv preprint arXiv:2410.22898, 2024.

[19] M. Hu, Z. Li, J. Yu, X. Wan, H. Tan, and Z. Lin, "Efficient-lightweight yolo: Improving small object detection in yolo for aerial images," Sensors, vol. 23, no. 14, p. 6423, 2023.

## AUTHORS

**Vaikunth M** is currently pursuing a Bachelor of Engineering at the College of Engineering, Guindy, Anna University, Chennai, India. He is actively engaged in computer vision research, specializing in object detection, segmentation, and classification, with a particular interest in exploring the segmentation of occluded objects in video streams. Vaikunth's professional experience spans the field of semiconductor electronics and Technology CAD coding, where he has conducted AC and DC analyses of various types of MOSFETs and implemented digital logic using 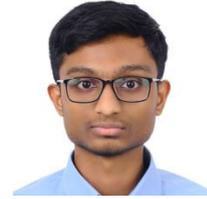 MOS systems. His technical skills are complemented by a strong inclination toward machine learning, particularly in its applications to the electronics industry to enhance computational efficiency. He has also pursued research projects in pure natural language processing. Additionally, Vaikunth has worked with retrieval-augmented generation systems and multi-agent models to harness the capabilities of large language models and generative AI. Vaikunth aspires to continue his research through future conferences and collaborations, contributing to advancements at the intersection of computer vision, machine learning, and generative AI.

**Dejey D** received her B.E. and M.E. degrees in Computer Science and Engineering from Manonmaniam Sundaranar University, Tirunelveli, India, in 2003 and 2005, respectively. Later, she was with the Department of Computer Science and Engineering, Manonmaniam Sundaranar University, Tirunelveli, India, as a Junior Research Fellow under the UGC Research Grant. She completed her Ph.D. in Computer Science and Engineering in 2011. She is an Associate Professor in the Department of Computer Science and Engineering, College of Engineering Guindy, 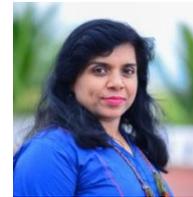 Anna University, Chennai. She had been with the Department of Computer Science and Engineering, Anna University Regional Campus – Tirunelveli, as an Assistant Professor from August 2010 to May 2023 and as the Head of the Department from 2011 to 2015. She is a member of IEEE, IEI, Indian Science Congress, and ISTE. She is the author of the book 'Cyber Forensics' that is published by Oxford University Press. She is the recipient of the Silver Award for Excellence in Informational Technology (IT) in AYUSH Sector from the Ministry of AYUSH, Government of India, and the Active Consultant Award from Anna University, Chennai. She holds two Indian Patents and two Indian Copyrights. She has more than 40 publications in reputed journals. She has published research papers at National and International Conferences. She has received funding from the Centre for International Relations, Anna University Chennai, India to present her research article at the International Conference in Melbourne, Australia in 2024. Her research interests include Signal, Image and Video Processing, IoT, and Multimedia Security.

**Vishaal C** is a final-year undergraduate student in Computer Science and Engineering at the College of Engineering, Guindy, Chennai. His academic and professional experiences demonstrate a strong passion for backend systems, enterprise-level solutions, and emerging technologies. He has worked extensively on integrating platforms like MongoDB and Splunk through efficient pipelines, ensuring scalability and reliability in enterprise environments. In addition to his backend 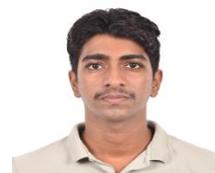 expertise, he possesses a keen interest in advanced fields such as computer vision, cybersecurity, and data analytics. His fascination with AI-driven technologies and their potential to transform industries drives his aspiration to pursue a Master's degree in Computer Vision. He is particularly motivated to address challenges involving visual data and contribute to groundbreaking research in this domain. Dedicated to innovation and problem-solving, he aims to make significant contributions to the global technology community as a researcher, developer and thought leader in his areas of interest.

**Balamurali S** is a Computer Science graduate student who is currently pursuing his B.E. in Computer Science and Engineering at the College of Engineering Guindy (CEG), Anna University. His research interests focus on how he can use technology to tackle challenges in sustainability, agriculture, and automation on the global level. which include applications in machine learning areas like predictive modeling to 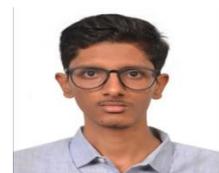 support personalized agricultural systems with tailored crop guidance and soil analysis for farmers, and also sustainable logistics systems that focus on emissions, integrate the use of real-time weather data, anomaly detection with optimized routing to reduce carbon footprints and enhance operational efficiency. His research in computer vision focuses on real-time object detection frameworks in applications ranging from safety enforcement to automation. He is motivated to drive innovative research that delivers impactful and practical solutions, paving the way for a sustainable and technologically advanced future.